\documentclass[letterpaper, 10 pt, conference]{ieeeconf}  

\IEEEoverridecommandlockouts                              

\usepackage{graphicx}
\usepackage{physics}
\usepackage{amsmath}
\usepackage{cite}
\usepackage{cleveref}
\usepackage{amsmath}

\usepackage{graphicx}
\usepackage{multirow}
\usepackage{booktabs}

\DeclareGraphicsExtensions{.pdf,.jpeg,.png}

\DeclareMathOperator*{\argmin}{arg\,min}
\Crefname{figure}{Fig.}{Figs.}

\makeatletter
\newcommand*{\rom}[1]{\expandafter\@slowromancap\romannumeral #1@}
\makeatother

\title{\LARGE \bf
Optimizing Base Placement of Surgical Robot:
Kinematics Data-Driven Approach by Analyzing Working Pattern
}

\author{Jeonghyeon Yoon$^*$, Junhyun Park$^*$, Hyojae Park, Hakyoon Lee, Sangwon Lee, and Minho Hwang$^{\dagger}$
\thanks{* Equal Contribution, ${\dagger}$ Corresponding author}
\thanks{This work was supported by the DGIST R\&D Program of the Ministry of Science and ICT (23-PCOE-02, 23-DPIC-20), by the DGIST Start-up Fund Program of the Ministry of Science and ICT (2024010213), and by the collaborative project with ROEN Surgical Inc.
This work was supported by the Korea Medical Device Development Fund grant funded by the Korea government (Project Number: 1711196477, RS-2023-00252244) and by the National Research Council of Science \& Technology (NST) grant funded by the Korea government (MSIT) (CRC23021-000).}
\thanks{Jeonghyeon Yoon, Junhyun Park, Hyojae Park, Hakyoon Lee, Sangwon Lee, and Minho Hwang are with the Department of Robotics and Mechatronics Engineering, DGIST, Daegu, 42988, Republic of Korea
        {\tt\small \{yjh1434, sean05071, hyojae, hakyun, ghkdlsxm275, minho\}@dgist.ac.kr}}%
}

\begin{document}

\maketitle
\thispagestyle{empty}
\pagestyle{empty}

\begin{abstract}

In robot-assisted minimally invasive surgery (RAMIS), optimal placement of the surgical robot base is crucial for successful surgery. Improper placement can hinder performance because of manipulator limitations and inaccessible workspaces. Conventional base placement relies on the experience of trained medical staff. This study proposes a novel method for determining the optimal base pose based on the surgeon’s working pattern. The proposed method analyzes recorded end-effector poses using a machine learning-based clustering technique to identify key positions and orientations preferred by the surgeon. We introduce two scoring metrics to address the joint limit and singularity issues: joint margin and manipulability scores. We then train a multi-layer perceptron regressor to predict the optimal base pose based on these scores. Evaluation in a simulated environment using the da Vinci Research Kit shows unique base pose score maps for four volunteers, highlighting the individuality of the working patterns. Results comparing with 20,000 randomly selected base poses suggest that the score obtained using the proposed method is 28.2\% higher than that obtained by random base placement. These results emphasize the need for operator-specific optimization during base placement in RAMIS.
\end{abstract}


\section{INTRODUCTION}
\setlength{\textfloatsep}{10.0pt plus 2.0pt minus 2.0pt}
\begin{figure}[!t]
    \centering
    \includegraphics[width=0.95\linewidth]{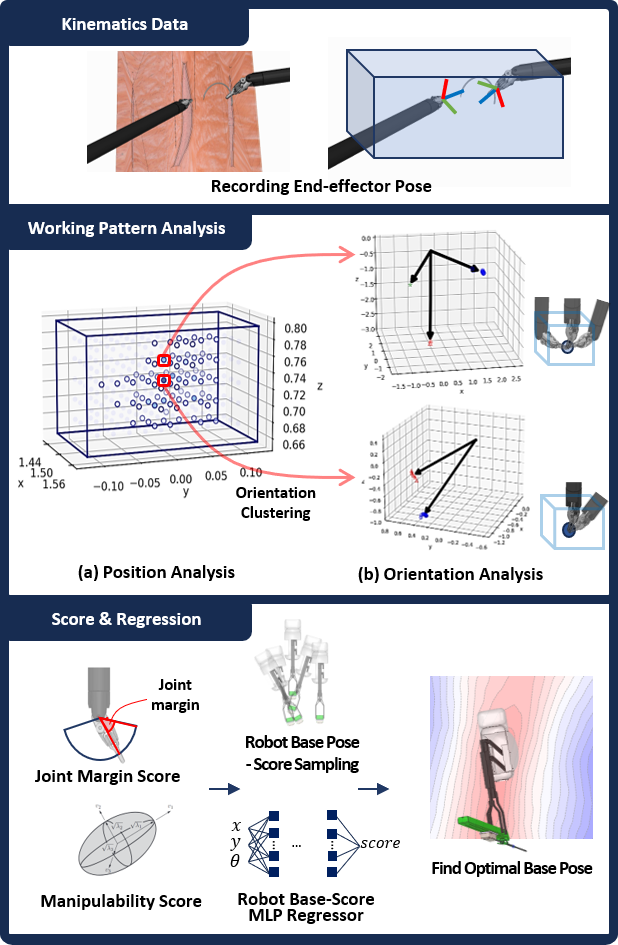}
    \caption{\textbf{Optimizing the robot base pose using the operator’s working pattern analysis. Kinematics Data:} To analyze the operator’s working pattern (e.g., needle handling and grasping strategies), we use the recorded end-effector pose data from the previous operation. \textbf{Working Pattern Analysis:} We can identify representative end-effector poses by analyzing the poses frequently adopted by the end-effector. \textbf{Score and Regression:} We define two kinematics metrics to calculate the scores for the base pose. We perform multilayer perceptron (MLP)-based regression to determine the optimal base pose.}
    \label{fig:intro}
\end{figure}
In manipulation tasks using robots, the base placement of the manipulator significantly affects key performance parameters, such as manipulability, reachability, and dexterity. In robot-assisted minimally invasive surgery (RAMIS), the placement of the surgical robot base or port is a preoperative procedure performed by trained medical staff. This procedure generally follows the broad guidelines provided by the robot company but relies mostly on their experience \cite{c1}. Similar to general tasks performed using a manipulator, this process in RAMIS becomes more crucial as improper base locations may lead to failure in operation because of manipulator constraints such as joint limits or singularities. This consequently causes an impossible configuration of the end-effector required for surgical tasks \cite{c2}. The problem becomes more evident when performing surgical tasks, such as suturing or anastomosis that require diverse changes in the orientation of the end-effector. The strategic placement of the surgical robot base significantly affects key factors, such as manipulability, reachability, and dexterity, which are critical for successful robot-assisted surgery \cite{c4}.

\begin{figure}[t!]
  \centering
  \includegraphics[width=\linewidth]{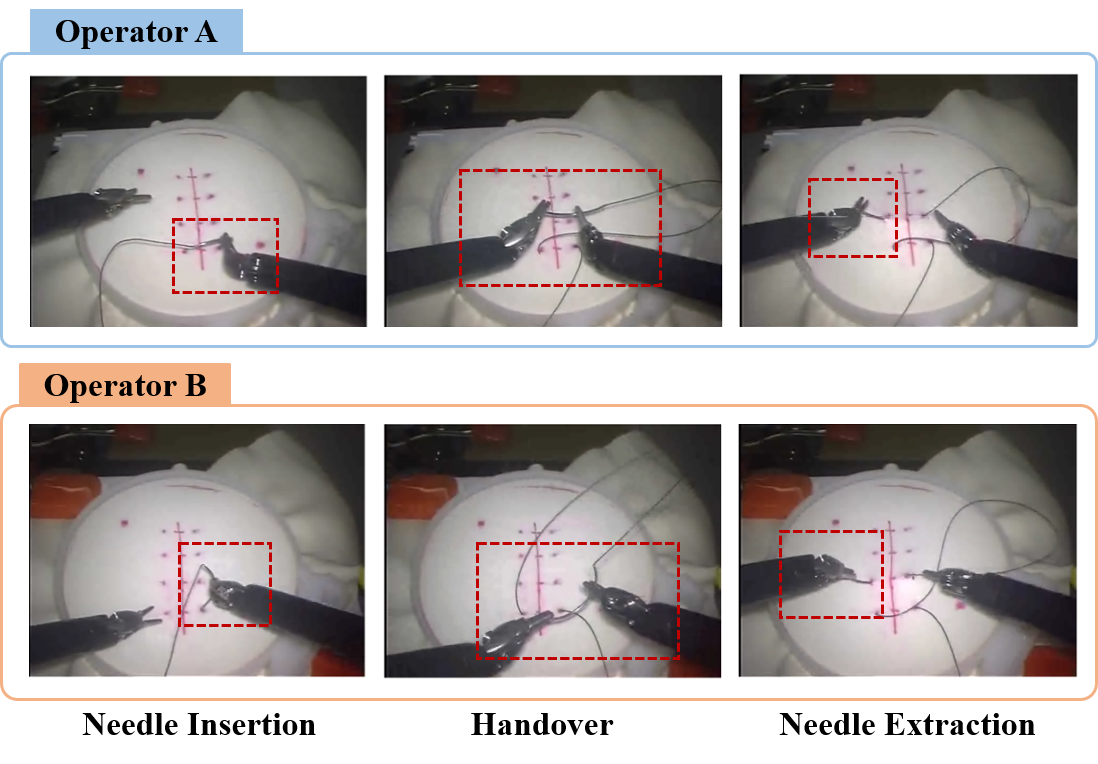}
  \caption{\textbf{Working Pattern Observation using the JIGSAWS dataset:} Among the JIGSAWS dataset \cite{jigsaw}, we selected two operators with expertise ($>$ 100 hrs) in operating surgical robots. Each operator exhibited different manipulation techniques when performing the suturing task (needle insertion, handover, needle extraction). These figures indicate that each operator has a unique working pattern.}
  \label{working_pattern}
\end{figure}

However, it is difficult to form a specific guideline for base placement because the proper base pose varies depending on the surgical task and the operator’s working pattern. The working pattern encompasses the operator’s preferred manipulation techniques and handling styles (including dominant hand usage, needle handling, and grasping strategies). These working patterns, illustrated in \Cref{working_pattern}, can vary significantly among operators. As described in \cite{c3}, finding the optimal base placement requires careful consideration of the operator’s working pattern, patient anatomy, and robot kinematics constraints.

By strategically positioning the robot base, operators gain improved access to the desired target with the robot end-effector, thereby reducing surgical duration and minimizing patient exposure to anesthesia. In situations where the base placement is not optimal, operators face the challenge of having to repeatedly adjust the end-effector pose to fulfill their tasks. Furthermore, each operator may have distinct working patterns associated with the end-effector poses used to reach the target or their preferred hand for specific surgical tasks. Considering these working patterns during base placement enables operators to operate within their accustomed workflows, thereby improving work efficiency.

This study proposes a novel method for optimizing the base pose of a robot by analyzing the operator’s working pattern. We analyze end-effector pose data to identify frequently visited positions and adopted orientations in the workspace, revealing insights into the established working pattern. The primary objective is to mitigate potential robot-related challenges within this pattern, such as joint limit and singularity problems. To achieve this, we introduce two distinct scoring metrics to evaluate the suitability of different base poses. A dataset of these scores and their corresponding base poses is then used to train a deep-learning regression model that predicts the optimal base pose for a given operator’s working pattern. The major contributions of this study can be summarized as follows:

\begin{itemize}
    \item We propose a novel kinematics data clustering approach that analyzes the operator’s working pattern in robot-assisted surgery by identifying key end-effector poses.
    \item Joint margin and manipulability scores are defined for the base pose to address joint limit and singularity concerns in frequently used end-effector poses by an operator.
    \item To determine the optimal robot base pose in continuous space, a trained MLP model is used for highly accurate base pose score regression.
\end{itemize}

\section{Related Work}
Various methods for determining the optimal base pose for surgical robots have been proposed to enhance surgical ergonomics and efficiency. Trejos et al., \cite{c5} identified potential port locations for cardiac surgery using global conditioning and isotropy indices, defining an optimal port but not addressing continuous base pose optimization. Sun and Yeung \cite{c6} extended this work by considering dexterity and reachability using the global isotropy index and efficiency index on continuous base pose space. Feng et al., \cite{feng_pose} maximized the robot’s operational workspace overlap with the surgical region, whereas Sundaram et al. \cite{Sundaram_task-specific} used robot capability maps to optimize the base pose and avoid task-specific reachability issues. However, these studies primarily relied on reachability and dexterity without incorporating operator kinematics for manipulability optimization.

Papanikolaidi et al., \cite{c7} used genetic algorithms for task prediction and manipulability optimization; however, their point-to-point assumptions limited real-world applications and relied on theoretical assumptions rather than real operator robot manipulation data. Other approaches optimized the base pose by minimizing collisions \cite{c9} or using patient specific organ displacement \cite{Maddah_vision_port}; however, they did not analyze operator-specific working patterns and use them for base pose optimization.

This study fills the research gap by proposing a novel approach that analyzes and uses operator-specific working patterns to determine an optimal base pose, maximizing both manipulability and work efficiency while considering the operating preferences.

\section{Methods}

The robot base, denoted as $(X, Y, \Theta)$, specifies the robot's horizontal position (represented by $X$ and $Y$) and its rotation around the vertical Z-axis (represented by $\Theta$). Notably, the vertical Z-axis translation is excluded from the base pose space because it is predetermined by the patient’s geometry, as detailed in reference \cite{c9}. Finally, we evaluate the proposed approach in the simulated environment using the da Vinci Surgical System.

We analyze past kinematics data of the operator to identify the positions and orientations that the end-effector frequently occupies during surgery, reflecting the operator’s working patterns. Furthermore, we consider joint limits and manipulability using a joint margin score and a manipulability score. By combining these scores, we create a dataset of potential base poses with corresponding scores. To bridge the gap between discrete samples and continuous space, we use an MLP-based data regressor trained on this dataset. This allows us to identify the base pose that maximizes the combined score, ultimately balancing operator comfort with robotic constraints.

\subsection{Operator Working Pattern Analysis}
\label{sec:data_anal}

The working patterns of the operator’s robot manipulation can be characterized by examining the frequent end-effector poses adopted during the surgical procedure, as described in the "Working Pattern Analysis" of \Cref{fig:intro}. In surgical procedures, such as needle threading or knot tying, orientation changes frequently occur. Furthermore, as the volume of the end-effector pose data increases, the computational complexity of determining the optimal base pose for all data becomes significant. From a computational cost perspective, it is more efficient to calculate scores for different base poses based on the end-effector poses representing the working pattern.

The working pattern analysis process consists of two sequential steps: 1) analyzing the position of the end-effector and 2) analyzing its orientation. For the initial step of position analysis, we divide the robot's workspace into small voxels. These voxels allow us to analyze the positional movements of the end-effector. Blue circles in (a)-Position Analysis represent the centers of these voxels, whereas the circles with borders highlight the specific voxels visited by the robot, and their intensity indicates the number of visits.

Next, we analyze the adopted orientations of the end-effector within the visited voxels, (b)-Orientation Analysis. To identify the most frequently used orientations within each visited voxel, we employ mean-shift clustering \cite{c10}. We use the rotation vector method to represent orientations to ensure compatibility with the Euclidean distance-based clustering algorithm. A parameter called bandwidth is carefully selected with silhouette score \cite{silhouettes} to ensure accurate cluster formation. Finally, we compute the average orientation within each cluster for every voxel, effectively capturing the most frequent end-effector poses within each region. Combining these representative orientations with the central positions of the voxels throughout the workspace produces a comprehensive set of end-effector poses that characterize the working pattern of the operator.

\subsection{Score Definition}
\label{sec:score_def}
Identifying the optimal base pose for the operator's working pattern requires ensuring frequently adopted end-effector poses maintain safe distances from joint limits and guarantee manipulability. Considering these factors, we define two metrics: joint margin and manipulability scores.

1) \textbf{Joint margin score ($score_{JM}$)}: This metric indicates how far the joint configuration for the representative end-effector pose is from the joint limits:
\begin{gather}
    q_{dist} = \abs{q_{pose} - q_{mid}}, \ \mu_{norm, dist} = \frac{q_{dist}}{q_{dist,  max}} \nonumber \\ 
    score_{JM} = \frac{\Sigma{(1 - \mu_{norm, dist})}}{\# \ joints}\\
    (0 \leq score_{JM} \leq 1) \nonumber
\end{gather}
The joint margin score can be determined by the closeness of the joint configuration ($q_{pose}$) to the central value of the allowable joint angles for each joint ($q_{mid}$). A shorter distance ($q_{dist}$) between the joint configuration and the central value implies a greater joint margin. $q_{dist}$ is calculated for each joint. This value is then normalized using min-max normalization to fall within the range of 0 and 1. Lower $\mu_{norm,dist}$ indicates a larger joint margin, it is subtracted from 1, and the result is divided by the number of joints ($\# \ joints$) to obtain the average joint margin. However, weighted average can be used by setting different weights for each joint (i.e. assigning higher weight to the joint frequently encountering its limit). The computed joint margin score ranges from 0 to 1.

2) \textbf{Manipulability score ($score_{M}$)}: This metric measures how far the robot is to singularity and its flexibility in moving. Manipulability was analyzed using a three-dimensional shape known as the manipulability ellipsoid. The eigenvalues ($\lambda$) of the matrix obtained by multiplying the Jacobian matrix (describing robot motion) and its transpose ($J \cdot J^{T}$) reveal the lengths of the principal axes of the manipulability ellipsoid. A value close to 1 for the ratio between the longest and shortest principal axes of the ellipsoid indicates a higher level of manipulability. Using this concept, we define the manipulability score as follows:
\begin{gather}
    \mu (J \cdot J^{T}) = \frac{\sqrt{\lambda_{\max} (A)}}{\sqrt{\lambda_{\min} (A)}} \cong \sqrt{\frac{\lambda_{\max} (A)}{\lambda_{\min} (A)}} \geq 1 \nonumber \\ 
    {score}_{LM} = \frac{1}{\mu_{\text{linear}}}, \quad score_{AM} = \frac{1}{\mu_{\text{angular}}} \nonumber \\
    (0 \leq score_{LM}, \  score_{AM} \leq 1) \nonumber \\
    score_{M} = score_{LM} + score_{AM}
\end{gather}

The Jacobian matrix is obtained by applying the chain rule to differentiate the transformation matrix, which describes the position and orientation of the robot’s end-effector relative to its base. Because this transformation involves both translation and rotation, the Jacobian matrix can be broken down into two parts: one representing linear velocity, and the other representing angular velocity. For each of these components, we determine the lengths of the longest and shortest principal axes of the manipulability ellipsoid, which are derived from the eigenvalues of the matrix obtained by multiplying each Jacobian matrix by its transpose. Subsequently, we compute the reciprocal of these ratios ($\mu_{\text{linear}},\ \mu_{\text{linear}}$) in a manner that assigns higher values closer to 1 a superior score.

Similar to the joint margin score, both linear and angular manipulability scores are bounded within the interval from 0 to 1. The final manipulability score ($score_{M}$) is the sum of the two manipulability scores ($score_{LM}$ and $score_{AM}$).

3) \textbf{Final score calculation}: For all representative end-effector poses, we calculate the joint margin score and manipulability score and then aggregate them to derive the final score for the base pose:
\begin{gather}\label{final_score}
    w_{voxel} =
    \begin{cases}
        1 \qquad \mathrm{(if \ visited)} \\
        \alpha \qquad \mathrm{(else)} \\
    \end{cases} \nonumber \\
    score_{final} = \sum\limits_{voxel}w_{voxel}(score_{JM} + score_{M})
\end{gather}

For visited voxels, we set the weight to 1, whereas for unvisited voxels, we introduced a parameter with values varying between 0 and 1. This parameter allows users to determine the importance assigned to unvisited voxels based on their preferences.

\subsection{Score Sampling and Regression}
\label{sec:regression}
\textbf{1) Score sampling}: 
We begin with a set of $N$ representative end-effector poses denoted by $\Sigma_{EE}$. Each pose is defined by the center position of the voxel in the workspace, $p_{EE_i}$, and a representative orientation in the rotation vector form, $\omega_{EE_i}$. From this set, we sample $M$ distinct robot base poses within a feasible range, forming the dataset $D$.
\begin{gather}
p_{EE_i} = (x_{voxel_{i}}, y_{voxel_{i}}, z_{voxel_{i}}) \nonumber \\
\omega_{EE_i} = (\omega_{x_i}, \omega_{y_i}, \omega_{z_i}) \nonumber \\
\Sigma_{EE} = \{p_{EE_i}, \omega_{EE_i}\}_{i=1}^{N} \\
D = \{(X,Y,\Theta)_{j}, score_{final_{j}}\}_{j=1}^M
\end{gather}

For each sampled base pose, we calculate the transformation matrix $T_{EE_i}^{base_j}$ between the base and the respective representative set of end-effector poses. We then solve the inverse kinematics to determine the corresponding joint angles. Using these joint angles, we calculate the $score_{JM}$ and $score_{M}$ as described in (1) and (2).
Finally, we combine these scores using (3) to obtain the $score_{final}$. This process generates one base pose-score pair. Repeating it for $M$ samples creates the dataset $D$ containing base pose parameters, $(X, Y, \Theta)_j$, and their corresponding $score_{final_{j}}$.

Score sampling relies heavily on inverse kinematics calculations. In this score sampling procedure, inverse kinematics calculations are necessary for every representative pose with respect to every sampled base pose($N \times M$ times). This highlights the significant computational demands of these tasks. For efficient calculations, we adopted closed-form inverse kinematics solutions introduced in \cite{c14} rather than numerical methods, which require significant computations.

\begin{figure}[t!]
  \centering
  \includegraphics[width=0.9\linewidth]{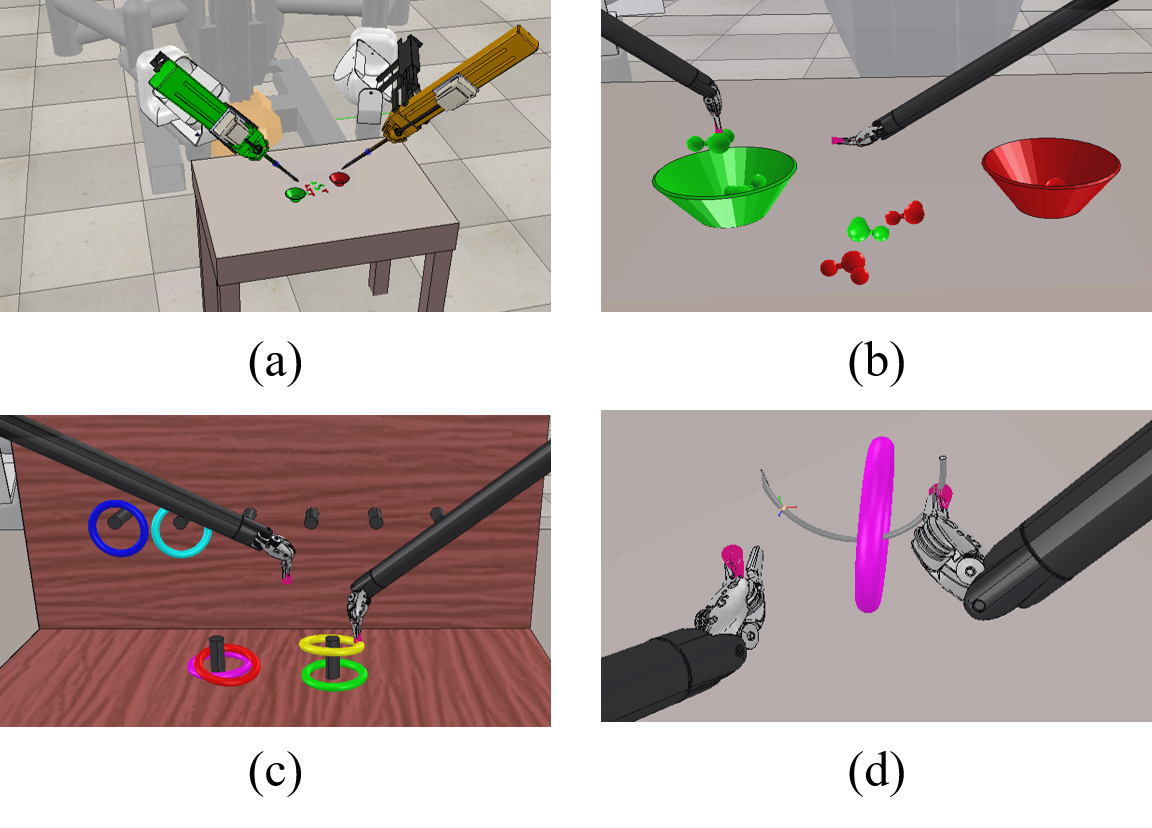}
  \caption{\textbf{Simulation setup}. (a) dVRK laparoscopic surgery training simulation setup. (b) Pick and place. (c) Peg on board. (d) Needle threading.}
  \label{simulation}
\end{figure}

\begin{figure*}[t!]
  \centering
  \includegraphics[width=\linewidth]{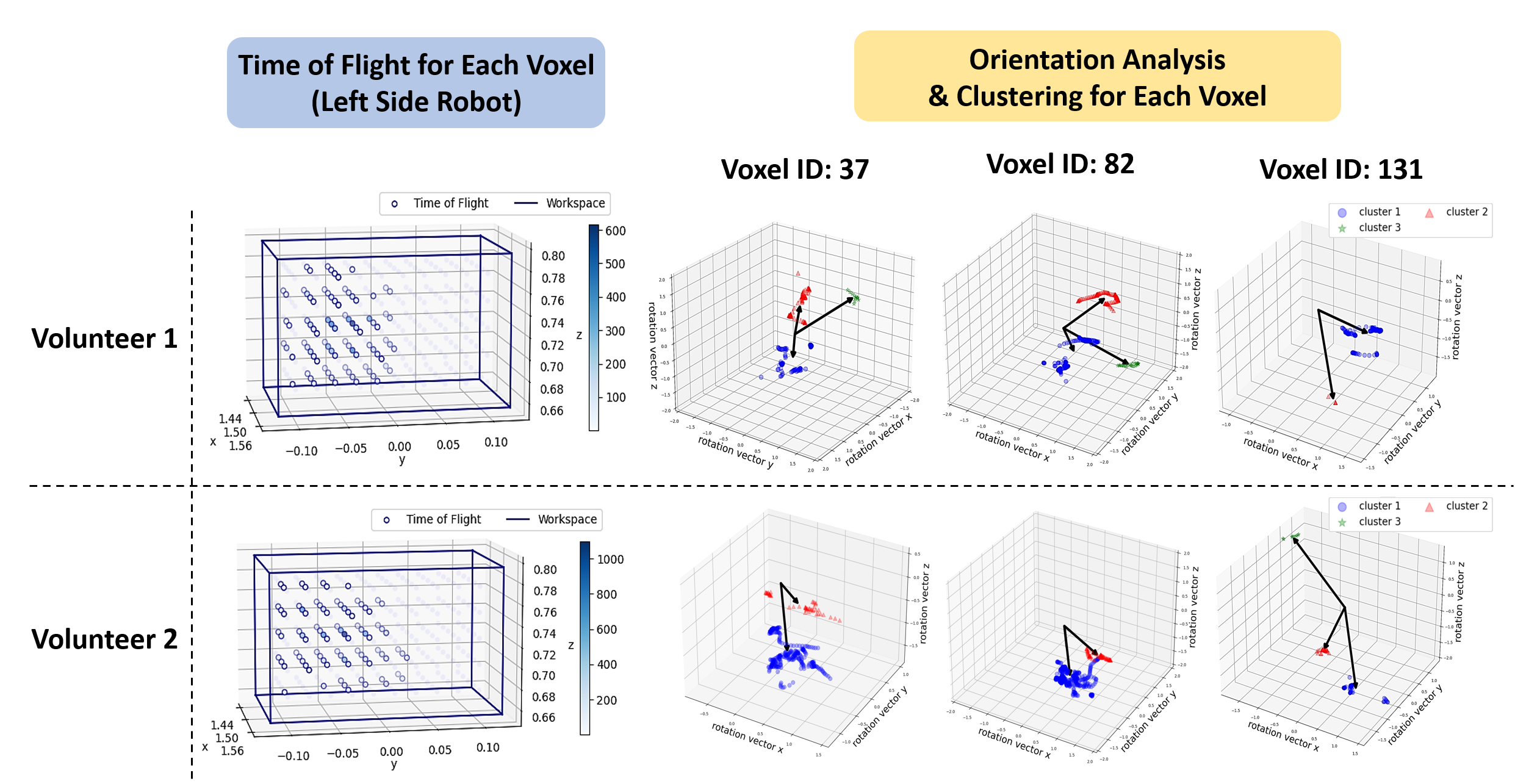}
  \caption{\textbf{Visited voxel and orientation clustering result}. The first column shows the scattered positions of the voxels visited by the end-effector during surgery. Sequential IDs are assigned to each voxel to analyze the working pattern. The remaining columns display the clustering orientation results for the commonly visited voxels. The black arrow indicates the representative orientations, which are the centroids of each orientation cluster, expressed as rotation vectors. Interestingly, each volunteer explored different sets of voxels and adopted his or her preferred orientations even within commonly visited areas.}
  \label{volunteer}
\end{figure*}

\textbf{2) Regression}: 
To effectively optimize the robot’s base pose, we seek a continuous representation of the base pose score space. This is achieved by training a regressor model on the collected dataset $D$. This model allows us to predict scores for any base pose within a feasible range, ultimately guiding us toward the globally optimal base pose with the highest score. 
\begin{align}
    \label{eq:regression}
    score_{final} = f_{\theta}(X_{norm}, Y_{norm}, \Theta_{norm})
\end{align}
This study explores three prominent regression models: the least absolute shrinkage and selection operator (LASSO) \cite{lasso}, support vector regressor (SVR) \cite{c11}, and MLP. We assessed their predictive performance by comparing their ability to accurately estimate final scores for unseen base poses. All input dimensions$(X, Y, \Theta)$ are normalized to the range [-1,1]. The regression model takes the normalized base pose as input and outputs the corresponding score as expressed in (6). The descriptions and settings for the three models are as follows.

\textbf{Lasso}: A linear regression that adds an L1-norm penalty to make it robust to outliers. The formula for LASSO is expressed as follows:
\begin{align}
    f(x) = \argmin_{w,b}\{ \frac{1}{n} \sum_{i=1}^{n}(y_{i} - \hat{y_{i}})^{2} + \alpha \sum_{j=1}^{m} \abs{w_{j}} \}
\end{align}

where $\hat{y}=w^{T}x+b$ represents the predicted value from the input with the weight vector of length $m$. $\alpha$ denotes a parameter indicating the degree of penalty, where a decrease in $\alpha$ leads to a smaller penalty for outliers. In this study, we set $\alpha$ to 0.5 for experimentation.

\textbf{SVR}: We use Gaussian Kernel SVR that maps data from the input space to a higher-dimensional space using a Gaussian function and finds a regression line that effectively describes the data. The formula for Kernel SVR is expressed as follows:
\begin{gather}
    f(x) = \sum_{i=1}^{n}(\alpha_{i}^{*} - \alpha_{i})K(x_{i}x_{j}) + b \nonumber \\
    K(x_{i}x_{j}) = \exp\{-\frac{\norm{x_{1} - x_{2}}_{2}^{2}}{2\sigma^{2}}\}
    \label{SVR}
\end{gather}

In equation \ref{SVR},  $\alpha_{i}$ and  $\alpha_{i}^{*}$ denote Lagrangian multipliers, and the detailed derivation can be found in reference \cite{c11}.

\textbf{MLP}: MLP is a feed-forward model that is composed of three hidden layers. Each layer consists of 48, 96, and 192 hidden units, respectively. We use the mean square error loss function, the adaptive moment estimation(Adam) optimizer with a learning rate of 0.0001 and 5000 epochs.

Using the trained regression model in this manner, we obtain a continuous score map for the robot base poses. Ultimately, using the best-performing model, we determine the optimal robot base pose $(X, Y, \Theta)$ that maximizes the score.

 \section{Simulation Setup and Data Collection}
To validate the proposed method, we noted a simulator designed for training surgeons in robotic skills, which uses the fundamentals of laparoscopic surgery training tasks  \cite{c12}. We created a simulated environment using the da Vinci Robot, where these tasks could be executed. We incorporated this environment using the robot simulator Coppeliasim, as detailed in reference \cite{c13}.

 \subsection{Simulation Setup}
 \label{sim_setup}
The environment established for training in reference \cite{c13} includes the ‘Pick and place’ and ‘Peg on board’ tasks (\Cref{simulation} (b), (c)). These two tasks involve lifting objects and placing them onto target objects. These tasks simulate motions during tissue dissection, where lifting tissue is required. They are characterized by frequent changes in position. In addition, we created a new environment for the needle threading task (\Cref{simulation} (d)). This task involves picking up a suturing needle and passing it through a ring. This task is characterized by frequent changes in orientation.

\subsection{Data Collection}
To ensure that volunteers had sufficient opportunity to refine their working patterns, we provided them with at least 50 hours of practice time. To collect data for training the regressors, we constructed three new environments with different target object placements for each of the three tasks mentioned in \Cref{sim_setup}, resulting in a total of nine environments. This allows volunteers to apply their working patterns developed during the practice phase in new environments. Data collection was performed while volunteers performed tasks in these nine environments. Each task took approximately 15 minutes, making the completion of all nine environments take approximately 135 minutes. On average, 18,839 end-effector pose data points were acquired for the entire set of tasks by each volunteer. The workspace dimensions were set to $0.2\times0.1\times0.1$[$m$], with a cubic voxel size of 0.02 meters. The size of workspace and voxel can be defined by surgical procedure and patient anatomy in real operation scenarios.

\setlength{\dbltextfloatsep}{2.0pt plus 2.0pt minus 2.0pt}

\section{Results}
To determine the optimal base pose with the proposed approach, we used end-effector poses recorded during the execution of tasks by four volunteers in a simulation environment. By analyzing their working patterns, we derived representative end-effector poses. These representative poses allowed us to calculate scores for various base poses, resulting in the creation of a dataset containing 20,000 base pose score pairs, $D$, through random selection of base poses. Subsequently, we trained regressors using the LASSO, SVR, and MLP methods to predict scores for base poses with $D$. We selected the model with the best learning performance among these regressors. Using this selected model, we identified the optimal base pose in the continuous base pose space with the maximum score. We validated its effectiveness by applying it to additional test tasks. The test environments consist of the three tasks illustrated in \Cref{simulation}, each with a new target object setup.

\begin{figure}[t!]
  \centering
  \includegraphics[width=\linewidth]{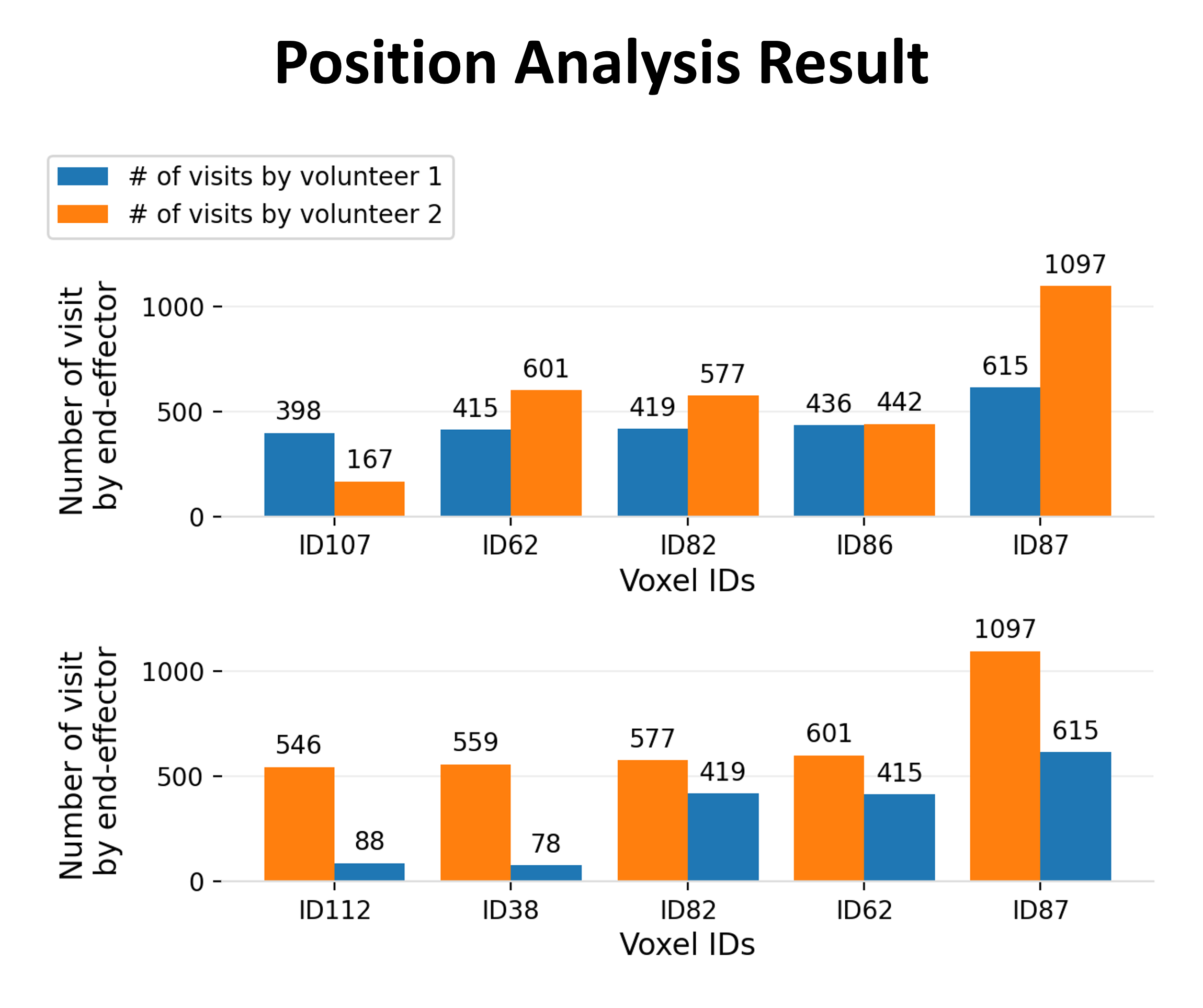}
  \caption{\textbf{Working pattern analysis : position analysis result}. This is the analysis result of the working pattern using the end-effector’s position for the two volunteers mentioned in Fig. 4. The graph in the first row compares the visit counts for the five voxels most frequently visited by volunteer 1, whereas the graph in the second row compares the visit counts for the five voxels most frequently visited by volunteer 2.}
  \label{position_result}
\end{figure}

\subsection{Working Pattern Comparison between Operators}
\label{Working Pattern Analysis}
Working pattern analysis of the end-effector pose dataset collected from the four volunteers revealed distinct patterns for each volunteer. \Cref{volunteer} presents the analysis results for two volunteers based on the left side robot’s end-effector pose data. The first column of \Cref{volunteer} shows the workspace divided into voxels, with each voxel colored according to the number of times the left robot’s end-effector visited it. The light blue circles represent the center positions of the voxels, and the bordered circles indicate the visited voxels. The darker the color inside the bordered circle, the more frequent the visits by the end-effector. For working pattern analysis, sequential IDs are assigned to each voxel.

The first column of \Cref{volunteer} indicates the position analysis of the working pattern, revealing distinct preferred working positions within the workspace. \Cref{position_result} compares visit counts for the five voxels where the end-effector of the left robot, operated by the two volunteers mentioned in  \Cref{volunteer}, stayed the longest. Analysis of the positions visited by the end-effector reveals that different working positions are preferred depending on the operator. These findings emphasize the individuality of working positions based on operator preference.

\begin{figure}[t!]
  \centering
  \includegraphics[width=\linewidth]{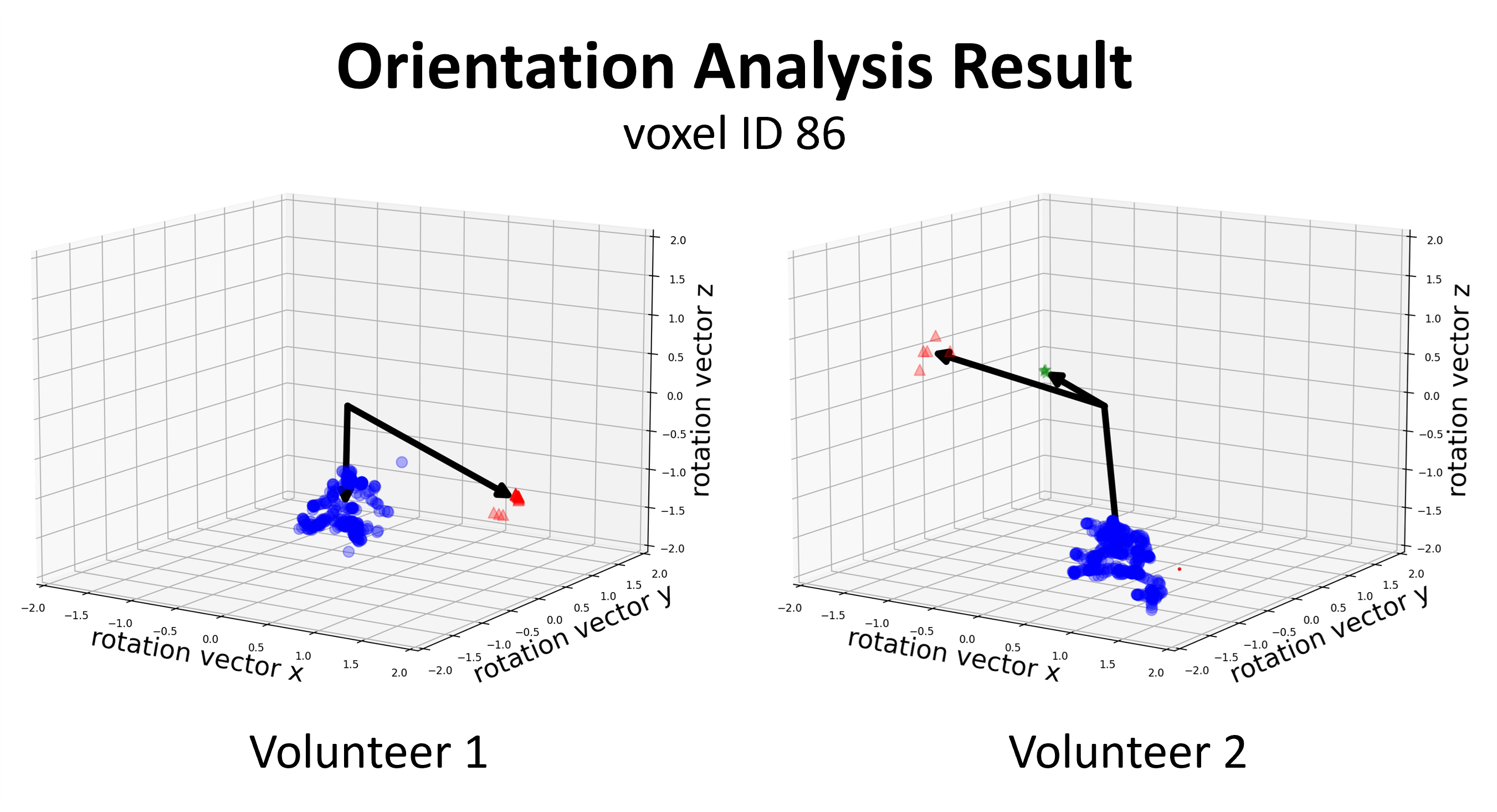}
  \caption{\textbf{Working pattern analysis : orientation analysis results}.The results show orientation clustering for voxels commonly visited by two volunteers with similar frequencies. For this voxel, the representative orientations for volunteer 1 are [0.0971, -0.188, -1.226] and [1.401, 0.711, -1.136], whereas, for volunteer 2, they are [0.434,-0.416,-1.746], [-0.191,-2.498,1.214], and [0.141,-1.203,0.734], indicating different outcomes (represented in rotation vector form).}
  \label{id86ori}
\end{figure}

\begin{figure*}[t!]
  \centering
  \includegraphics[width=\linewidth]{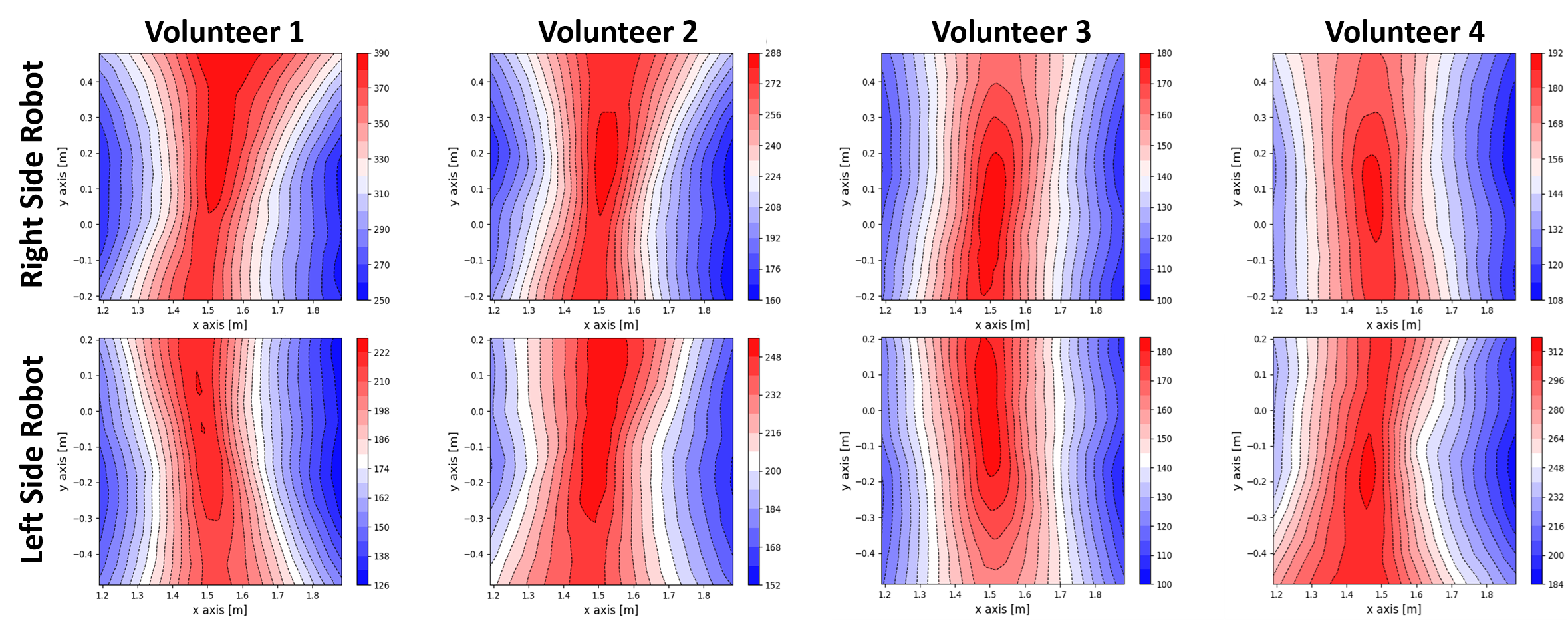}
  \caption{\textbf{Base pose score contour map.}  The scores are expressed as contour maps in the base pose of the four volunteers. For visibility, $\Theta$ is fixed at  $\Theta$ is fixed at -90 $^\circ$, and the distribution of scores according to changes in $X$ and $Y$ can be observed. Each column represents volunteers 1$-$4, and each row represents the base pose score contour map of the right and left robots.}
  \label{contour}
\end{figure*}

\begin{table}[t!]
\centering
\caption{The Performance Comparisons Between Regressors}
\label{The Performance Comparisons Between Regressors}

\begin{tabular}{ccccc} 
\toprule
\multicolumn{2}{c}{}            & LASSO     &~~~~ SVR& MLP                                                                                                                                                                                                                                    \\ 
\midrule
\multirow{2}{*}{~ $b_R \textrm{    score}$~~} & ~ RMSE~~ & ~ 24.00~~ &~~~~ 2.38& \textbf{0.32}                                                                                                                                                                                                                              \\
                       & SD       & 28.02     &~~~~ 3.37& \textbf{0.40}                                                                                                                                                                                                                              \\ 
\midrule
\multirow{2}{*}{$b_L\textrm{    score}$}     & RMSE     & 25.21     &~~~~ 2.20& \textbf{0.32}    \\
                       & SD       & 25.21     &~~~~ 3.08& \textbf{0.40}                                                                                                                                                                                                                              \\ 
\midrule
\multicolumn{5}{l}{\begin{tabular}[c]{@{}l@{}}\footnotesize * $b_R$, $b_L$ presents left and right robot base poses, respectively.\\ * RMSE and SD represent the root mean square error and standard \\ deviation between the ground truth and model output, respectively.\end{tabular}}   
\end{tabular}
\end{table}

\begin{table*}
\centering
\caption{Test Task Base Pose Score Comparison between the Proposed Method and Random Base Pose Selection}
\label{test task}
\begin{tabular}{ccccccccc} 
\toprule
\multirow{2}{*}{\begin{tabular}[c]{@{}c@{}}~\textbf{Test Task Base} \\\textbf{Pose Score}\\\text{[score]}\end{tabular}} & \multicolumn{2}{c}{Volunteer 1} & \multicolumn{2}{c}{Volunteer 2} & \multicolumn{2}{c}{Volunteer 3} & \multicolumn{2}{c}{Volunteer 4}  \\ 
\cmidrule{2-9}
                                                                                      & $b_R$          & $b_L$           & $b_R$          & $b_L$           & $b_R$          & $b_L$           & $b_R$          & $b_L$            \\ 
\midrule
Proposed
  Method                                                                     & \textbf{232.2} & \textbf{200.2} & \textbf{180.1} & \textbf{224.6} & \textbf{211.2} & \textbf{235.7} & \textbf{155.5} & \textbf{229.7}  \\ 
\midrule
Random
  Selection                                                                    & 155.3 ± 20.7     & 178.4 ± 25.2      & 107.3 ± 17.2     & 198.5 ± 27.8     & 142.2 ± 20.7     & 125.4 ± 17.5     & 102.5 ± 16.9     & 188.9 ± 24.4      \\
\midrule
\multicolumn{8}{l}{\begin{tabular}[c]{@{}l@{}}\footnotesize * $b_R$ and $b_L$ denote the left and right robot base poses, respectively.\end{tabular}}

\end{tabular}
\end{table*}

End-effector orientation analysis further reveals unique patterns in commonly visited voxels. The last three columns of \Cref{volunteer} illustrate the analysis of end-effector orientations adopted by the left robots operated by two volunteers in the commonly visited voxels. Each scatter plot, represented by different shapes, shows the end-effector orientations expressed in the rotation vector form after mean-shift clustering. The black arrows indicate the centroids of each cluster, representing the representative orientation at each voxel. Notably, even in the voxels visited by the end-effector operated by both volunteers simultaneously, distinct end-effector orientations are observed. In \Cref{position_result}, voxel ID 86 was visited by both volunteers simultaneously, and the visit counts were also similar, with 436 and 442 visits, respectively. The results of the derivation of the representative orientation for each volunteer are shown in \Cref{id86ori}. Despite both volunteers visiting simultaneously with similar frequencies, the orientation clusters are distributed differently, ultimately resulting in different representative end-effector poses.

\subsection{Regressor Evaluation}

We randomly set the base pose of the robot and sampled 20,000 scores for each base pose as described in \Cref{sec:regression}. The range of the right robot base pose space is $X$[$m$]= [1.188, 1.888], $Y$[$m$]= [-0.212, 0.488], $\Theta$[$^\circ$]= [-120,  -60] and the range of the left robot base pose space is $X$[$m$]= [1.190, 1.890], $Y$[$m$]= [-0.487, 0.213], $\Theta$[$^\circ$]= [-120, -60]. 

We performed the regression analysis outlined in \Cref{sec:regression} using the dataset $D$. Table \rom{1} presents the performance assessments, calculating the average RMSE and standard deviation(SD) against ground truth data for four volunteers. Remarkably, the MLP model excelled, achieving an RMSE of 0.32 and SD of 0.40 for both the right and left robot base pose-score regressors. These results highlight our precision-focused score regressor compared to with the SVR method of Xu et al., \cite{c9}. The superior performance of MLP is due to its adeptness at extracting complex features of the base pose-score dataset. As MLP excels in efficiently capturing and leveraging complex features, it outperforms SVR and Lasso in accurately modeling the underlying complexities of the dataset.

\subsection{Validation of Optimal Base Pose}
\Cref{contour} shows a contour map that visualizes score distributions across the continuous base pose space, generated using trained MLP regressors specific to each volunteer. This graphical representation distinctly shows that each volunteer exhibits a unique base pose score distribution, underlining variations in their working patterns, including their frequently adopted end-effector poses. Remarkably, our results indicated that each of the four volunteers had a unique optimal base pose, emphasizing the individuality of their working patterns.

To determine the optimal base pose with the highest global score, we use the MLP regressor for left and right robot base pose scores, incorporating 512,000 base poses as inputs. These poses are systematically spaced at intervals of $0.005 m$ for $X$, $0.005 m$ for $Y$, and $0.5 ^\circ$ for $\Theta$. We then identify the optimal base pose that produces the highest global score.

To evaluate the effectiveness of these optimal poses, we evaluated their performance on unseen end-effector data from different test tasks (pick and place, peg-in-hole, needle threading). We compared the scores obtained using the optimal base poses with those from 1,000 random base placements.

Table \rom{2} summarizes the results. The "Proposed Method" row shows scores obtained using the optimal poses, and the "Random Selection" row shows the average score and SD from random placements. For all volunteers, the proposed method significantly outperformed random placements across all test tasks.

These results suggest the effectiveness of our optimal base poses, which is probably due to similarities in working patterns between the training and test tasks. However, relying solely on our collected training data has limitations. Volunteers occasionally explore new areas and adopt different poses in test tasks not captured in the training dataset. Incorporating a more diverse dataset can yield more adaptable and robust optimal base poses that better reflect real-world surgical scenarios.

\section{CONCLUSIONS}

We proposed an approach that optimizes robot base poses by analyzing the working patterns of the operator, identifying key end-effector poses, and calculating scores using joint margins and manipulability. To determine the optimal robot base pose within a continuous space, we use a trained MLP regressor that takes the base pose as input and produces a corresponding score as output.

In a simulated environment, we collected end-effector pose data from four volunteers performing tasks, which revealed variations in working patterns and score distributions for base poses. Three test tasks further validated the proposed method, showing that scores at optimal base poses derived using the proposed method consistently outperformed those derived from random sampling.
 
This study addresses the challenges in achieving familiar end-effector poses during RAMIS caused by joint limits and singularities. The proposed approach is adaptable to additional score definitions. The use of an MLP-based regressor ensures improved accuracy, even with diverse scores. Future research will extend this study by defining additional scores for comprehensive consideration in RAMIS.

\addtolength{\textheight}{-12cm}   

\bibliographystyle{IEEEtran}
\bibliography{ref}

\end{document}